\documentclass[conference]{IEEEtran}
\IEEEoverridecommandlockouts
\usepackage{cite}
\usepackage{amsmath,amssymb,amsfonts}
\usepackage{algorithm}
\usepackage{algorithmic}
\usepackage{graphicx}
\usepackage{textcomp}
\usepackage{xcolor}
\usepackage{comment}
\usepackage{multirow}
\usepackage{pifont}
\usepackage{booktabs}
\usepackage{makecell}
\usepackage{caption}
\newcommand{\cmark}{\ding{51}} 

\def\BibTeX{{\rm B\kern-.05em{\sc i\kern-.025em b}\kern-.08em
    T\kern-.1667em\lower.7ex\hbox{E}\kern-.125emX}}
\begin{document}

\title{Uncertainty-Guided Inference-Time Depth Adaptation for Transformer-Based Visual Tracking}


\author{%
\IEEEauthorblockN{%
\setlength{\tabcolsep}{0pt}%
\begin{tabular*}{\textwidth}{@{\extracolsep{\fill}}cccc@{}}
Patrick Poggi\IEEEauthorrefmark{1} &
Divake Kumar\IEEEauthorrefmark{1} &
Theja Tulabandhula\IEEEauthorrefmark{1} &
Amit Ranjan Trivedi\IEEEauthorrefmark{1}
\end{tabular*}%
}
\IEEEauthorblockA{%
\IEEEauthorrefmark{1}University of Illinois at Chicago\\
{\ttfamily\small \{ppogg, dkumar33, theja, amitrt\}@uic.edu}%
}
}

\maketitle

\begin{abstract}
Transformer-based single-object trackers achieve state-of-the-art accuracy but rely on fixed-depth inference, executing the full encoder--decoder stack for every frame regardless of visual complexity, thereby incurring unnecessary computational cost in long video sequences dominated by temporally coherent frames. We propose UncL-STARK, an architecture-preserving approach that enables dynamic, uncertainty-aware depth adaptation in transformer-based trackers without modifying the underlying network or adding auxiliary heads. The model is fine-tuned to retain predictive robustness at multiple intermediate depths using random-depth training with knowledge distillation, thus enabling safe inference-time truncation. At runtime, we derive a lightweight uncertainty estimate directly from the model's corner localization heatmaps and use it in a feedback-driven policy that selects the encoder and decoder depth for the next frame based on the prediction confidence by exploiting temporal coherence in video. Extensive experiments on GOT-10k and LaSOT demonstrate up to 12\% GFLOPs reduction, 8.9\% latency reduction, and 10.8\% energy savings while maintaining tracking accuracy within 0.2\% of the full-depth baseline across both short-term and long-term sequences.
\end{abstract}


\section{Introduction}

\begin{figure*}[ht!]
    \centering
    \includegraphics[width=0.8\textwidth]{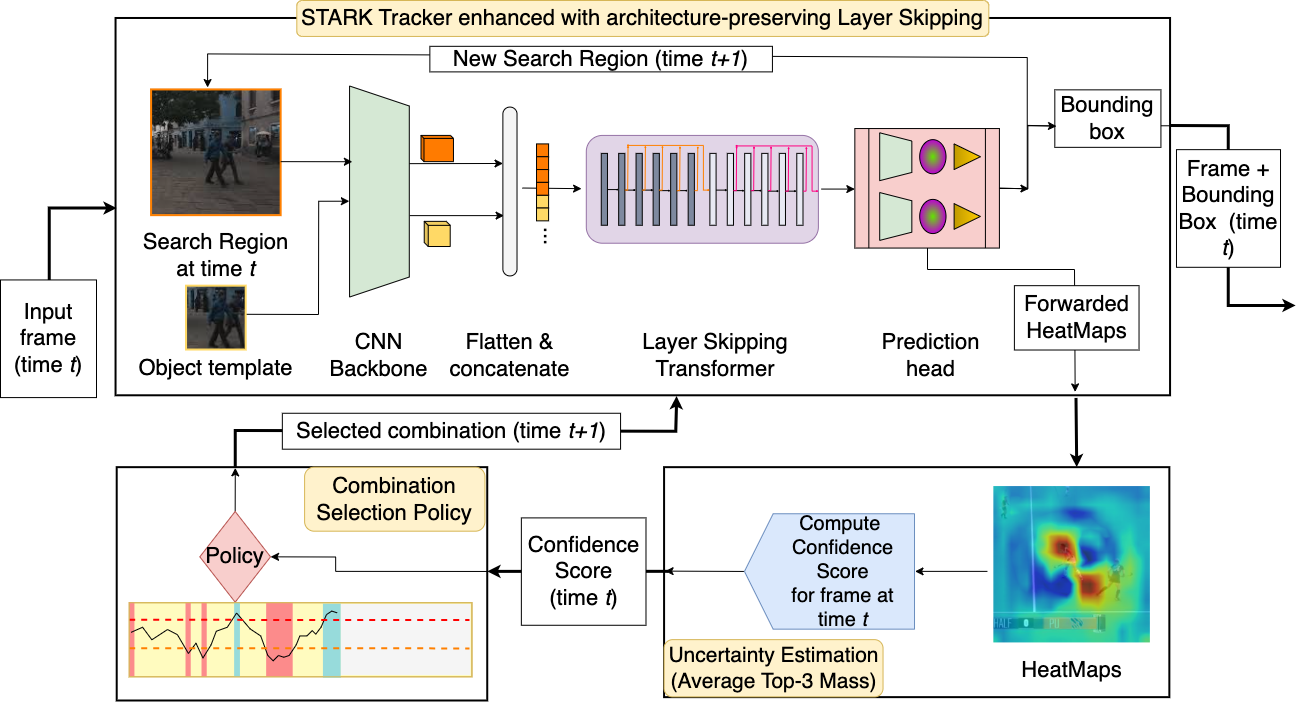}
\caption{Overview of the UncL-STARK framework. Uncertainty derived from corner localization heatmaps at frame $t$ drives inference-time depth adaptation by selecting the encoder--decoder depth for frame $t+1$. This feedback mechanism exploits temporal coherence in video to reduce computation while preserving tracking accuracy.}
    \label{fig:complete_framework}
\end{figure*}

Transformer-based architectures have set new performance benchmarks in single object tracking (SOT), enabling robust localization under occlusion, background clutter, illumination changes, and target deformation. Trackers such as STARK~\cite{yan2021stark}, TransT~\cite{chen2021transt}, and MixFormer~\cite{cui2022mixformer} achieve this performance using deep encoder--decoder stacks with multi-head self- and cross-attention~\cite{vaswani2017attention} to model long-range spatial dependencies and template--search interactions. However, this accuracy comes at a high computational cost: the full transformer depth is executed for every frame, irrespective of visual complexity or temporal coherence. In practice, most tracking sequences are dominated by visually simple, temporally stable frames with minimal inter-frame variation, making uniform full-depth inference highly redundant, particularly in long-duration or resource-constrained deployments.

Dynamic neural networks address this inefficiency by adapting computation to input difficulty via early exiting, adaptive depth, or token pruning~\cite{teerapittayanon2016branchynet,fan2019layerdrop,elbayad2019depth,rao2021dynamicvit,liang2022not}. While extensively studied in image classification and increasingly explored in object detection, dynamic computation remains underdeveloped in visual tracking. Existing dynamic trackers typically introduce architectural modifications, such as auxiliary prediction heads or learned gating networks, which increase model complexity, add parameters, and complicate training and deployment. Moreover, many rely on heuristic confidence thresholds or task-specific controllers rather than principled signals for regulating computation.

A key missing component in dynamic tracking systems is reliable, low-overhead uncertainty estimation. Uncertainty naturally guides adaptive computation: confident predictions require less processing, while uncertain cases benefit from deeper reasoning. Classical approaches such as deep ensembles~\cite{lakshminarayanan2017simple} and Monte Carlo dropout~\cite{gal2016dropout} provide well-calibrated uncertainty but require multiple forward passes, making them impractical for real-time tracking. By contrast, modern trackers already produce dense spatial predictions, such as corner or centerness heatmaps, whose spatial concentration implicitly encodes localization confidence. Sharp, peaked heatmaps indicate high certainty, while diffuse responses reflect ambiguity or occlusion. Although this signal has been exploited in object detection~\cite{law2018cornernet,zhou2019objects,gasperini2021certainnet}, it remains largely unused for adaptive computation in tracking.

This work bridges dynamic computation and lightweight uncertainty estimation to enable efficient, adaptive transformer-based tracking. We propose UncL-STARK, an architecture-preserving framework that performs inference-time depth adaptation by deriving uncertainty directly from corner localization heatmaps. The tracker is fine-tuned using random-depth training with knowledge distillation to remain predictive at multiple intermediate depths, enabling safe depth truncation without modifying the network or adding auxiliary heads. At runtime, a scalar confidence score computed from the top-$k$ heatmap mass drives a feedback policy that selects the encoder and decoder depth for the next frame, explicitly exploiting temporal coherence across video frames.

Our contributions are threefold. \textbf{First}, we introduce an architecture-preserving, depth-adaptive inference strategy for transformer-based tracking that enables selective execution of encoder and decoder layers without structural modification. \textbf{Second}, we propose a lightweight, heatmap-derived uncertainty proxy suitable for real-time tracking. \textbf{Third}, we develop a feedback-driven depth selection policy that achieves up to 12\% GFLOPs reduction, 8.9\% latency improvement, and 10.8\% energy savings while maintaining tracking accuracy within 0.2\% of the full-depth baseline on GOT-10k and LaSOT. Extensive experiments demonstrate that the approach generalizes across sequence lengths and difficulty levels, yielding a favorable efficiency--accuracy trade-off through principled uncertainty-guided adaptation.

\section{Related Work}
\label{sec:related_work}

\subsection{Transformer-Based Visual Tracking}

Single object tracking has evolved from correlation-based Siamese architectures~\cite{bertinetto2016siamfc,li2019siamrpn++,wang2019siammask,guo2020siamcar} to transformer-based models leveraging self- and cross-attention for template--search interaction. Following vision transformers~\cite{dosovitskiy2021vit}, TransT~\cite{chen2021transt} introduced attention-based fusion, and STARK~\cite{yan2021stark} adopted an encoder--decoder architecture with object queries and corner heatmap prediction. While effective, these trackers execute the full transformer depth for every frame. Efficiency-oriented work largely relies on offline architectural design and compression. MixFormer~\cite{cui2022mixformer} and MixFormerV2~\cite{cui2023mixformerv2} unify tokens within a single backbone and reduce depth via architectural simplification and distillation, while other approaches~\cite{ye2022joint} explore unified feature learning. However, these methods yield fixed-depth models with constant per-frame computation. In contrast, our approach enables dynamic, per-frame depth adaptation within a single trained model.

\subsection{Dynamic Neural Networks}

Dynamic neural networks reduce inference cost by adapting computation to input difficulty. Early-exit methods attach prediction heads at intermediate depths~\cite{teerapittayanon2016branchynet,huang2018msdnet,kaya2019shallowdeep}, but introduce auxiliary parameters and require careful loss balancing. For transformers, dynamic computation has been explored via token pruning and depth adaptation, including DynamicViT~\cite{rao2021dynamicvit}, LayerDrop~\cite{fan2019layerdrop}, and depth-adaptive transformers~\cite{elbayad2019depth}. These techniques have shown promise mainly in classification.

Dynamic computation in tracking remains limited. DyTrack~\cite{zhu2024dytrack} introduces instance-specific early exiting using auxiliary decision networks and prediction heads. In contrast, we preserve the original tracker architecture, introduce no auxiliary heads or gating modules, and enable depth adaptivity via random-depth training with knowledge distillation, guided by a lightweight uncertainty signal rather than learned controllers.

\subsection{Uncertainty Estimation for Vision}

Uncertainty estimation is commonly addressed using deep ensembles~\cite{lakshminarayanan2017simple} or Monte Carlo dropout~\cite{gal2016dropout}, but these require multiple forward passes and are impractical for real-time tracking. For localization, dense prediction maps provide implicit uncertainty cues: CornerNet~\cite{law2018cornernet} and CenterNet~\cite{zhou2019objects} show that heatmap magnitude and concentration correlate with confidence, and CertainNet~\cite{gasperini2021certainnet} derives sampling-free uncertainty directly from detection heatmaps. Conformal prediction on vision and robotics has been explored to extract uncertainty estimates with minimal compute overheads~\cite{kumar2025calibrated,kumar2025learnable,kumar2025uncertainty,stutts2023lightweight,stutts2024mutual}.

Despite this, uncertainty estimation is underutilized in tracking. UAST~\cite{zhang2022uast} and UncTrack~\cite{yao2025unctrack} introduce explicit uncertainty prediction branches or losses, increasing model complexity. In contrast, we derive uncertainty directly from the corner heatmaps already produced by the tracker, without additional outputs or training objectives. A scalar confidence score computed via top-$k$ probability mass after spatial softmax is then used to drive feedback-based depth adaptation, directly coupling uncertainty to inference-time computation.

\begin{figure}[t]
    \centering
    \includegraphics[width=\linewidth]{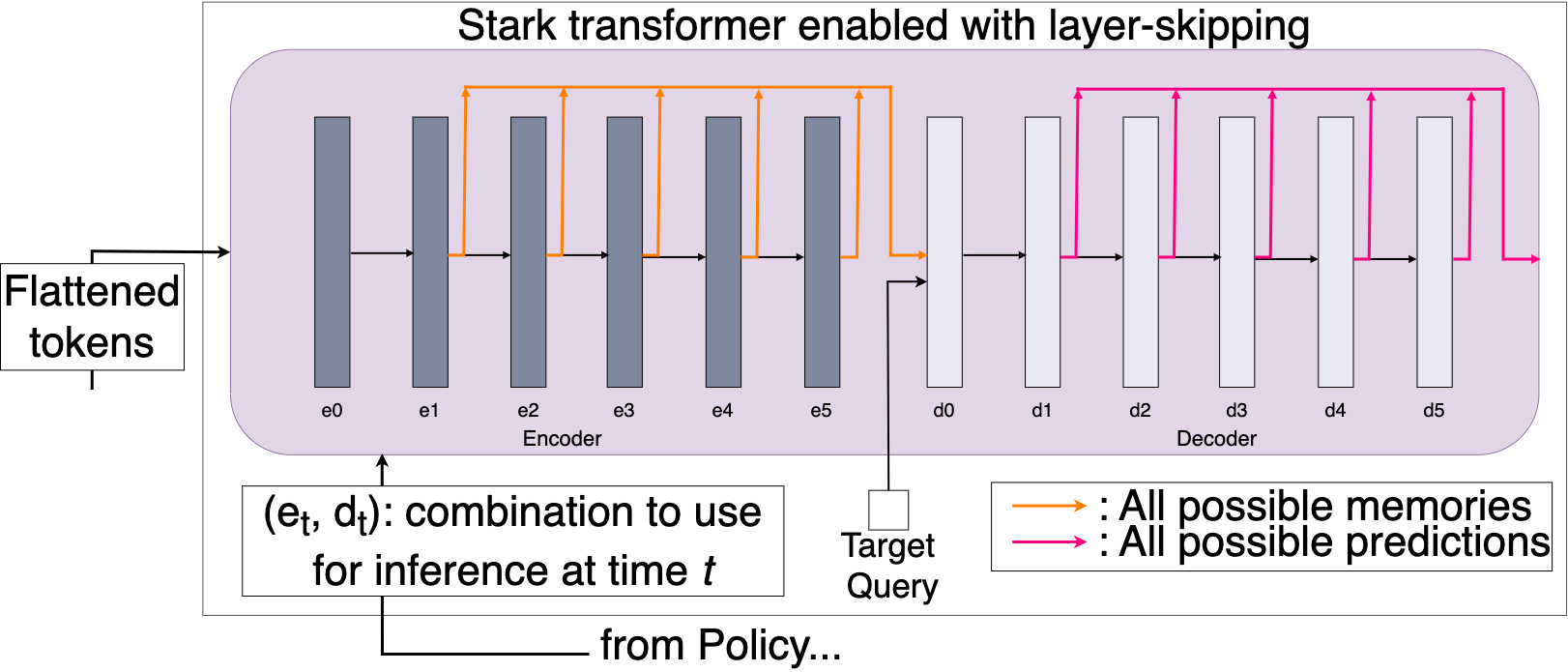}
\caption{Architecture-preserving depth truncation in UncL-STARK. The decoder attends to the output of a selected encoder layer, and predictions are produced from a selected decoder layer, enabling inference at arbitrary depths without modifying the prediction head.}
    \label{fig:LS_enab_transf}
\end{figure}

\section{Uncertainty-Guided Inference-Time Prediction Model Depth Adaptation}

We propose \emph{UncL-STARK}, an adaptive transformer-based single object tracking framework that enables inference-time depth adaptation using lightweight uncertainty estimates derived from corner localization heatmaps. The method strictly preserves the original STARK architecture and prediction head, introducing no auxiliary branches, gating modules, or additional parameters. Depth adaptivity is enabled by (i) exposing intermediate encoder and decoder layers for controlled truncation and (ii) fine-tuning the model with random-depth sampling and knowledge distillation to ensure reliable predictions at intermediate depths. At runtime, the tracker, uncertainty estimator, and depth selection policy operate in a feedback loop: heatmaps produced at frame $t$ yield a confidence score that determines the encoder--decoder depth for frame $t+1$ by exploiting temporal coherence to allocate computation where needed.

\subsection{Architecture-Preserving Model Depth Truncation}

UncL-STARK builds on the original STARK encoder--decoder transformer by exposing both encoder and decoder layers as depth-selectable components. For each frame $t$, inference is performed using a selected depth pair $(E_t, D_t)$ corresponding to the last executed encoder and decoder layers. Layers are $0$-indexed; selecting $(E_t, D_t)$ executes the encoder up to layer $E_t$ and the decoder up to layer $D_t$. The full-depth configuration $(N_{enc}-1, N_{dec}-1)$ exactly reproduces standard STARK behavior, while shallower configurations reduce computation by truncating deeper layers.

Formally, the encoder consists of $N_{enc}$ transformer layers
$\mathcal{E}_0,\dots,\mathcal{E}_{N_{enc}-1}$, updating hidden states as
\[
\mathbf{h}_{\ell+1} = \mathcal{E}_{\ell}(\mathbf{h}_{\ell}),
\qquad \ell = 0,\dots,N_{enc}-1.
\]
Standard STARK executes all layers and outputs
\[
\mathbf{m}_{\text{full}} = \mathbf{h}_{N_{enc}-1}.
\]
In UncL-STARK, an encoder depth parameter $E$ defines the memory
\[
\mathbf{m}^{(E)} = \mathbf{h}_{E},
\qquad E \in \{\mathrm{MIN}_{enc},\dots,N_{enc}-1\},
\]
corresponding to truncation at depth $E$ without architectural modification. The decoder comprises $N_{dec}$ layers
$\mathcal{D}_0,\dots,\mathcal{D}_{N_{dec}-1}$ operating on query representations
$\mathbf{q}_{\ell} \in \mathbb{R}^{N_q \times d}$. Given encoder memory $\mathbf{m}^{(E)}$,
\[
\mathbf{q}_{\ell+1} = \mathcal{D}_{\ell}(\mathbf{q}_{\ell};\, \mathbf{m}^{(E)}),
\qquad \ell = 0,\dots,N_{dec}-1,
\]
with $\mathbf{q}_0=\mathbf{0}$. While STARK outputs $\mathbf{q}_{N_{dec}}$, UncL-STARK allows early exit via
\[
\mathbf{q}^{(D)} = \mathbf{q}_{D},
\qquad D \in \{\mathrm{MIN}_{dec},\dots,N_{dec}-1\}.
\]

All encoder and decoder layers preserve identical input--output interfaces, and the prediction head remains unchanged. Representations $\mathbf{m}^{(E)}$ and $\mathbf{q}^{(D)}$ are therefore fully compatible with the original head, ensuring strict architecture preservation. Minimum depths $\mathrm{MIN}_{enc}$ and $\mathrm{MIN}_{dec}$ ensure sufficient feature abstraction before truncation.

\subsection{Training for Multi-Depth Inference}
\begin{algorithm}[t]
\caption{Random-Depth Fine-Tuning with Teacher--Student Distillation}
\label{alg:finetune_teacher_student}
\small
\begin{algorithmic}[1]
\REQUIRE Pretrained tracker $f_\theta$, dataset $\mathcal{D}$
\REQUIRE Max depths $(N_E,N_D)$, min depths $(E_{\min},D_{\min})$
\REQUIRE Distillation weight $\lambda$, learning rate $\eta$
\ENSURE Parameters $\theta^\star$ robust to truncated depths

\STATE Freeze backbone parameters
\STATE Teacher depth $(E_T,D_T)\leftarrow(N_E,N_D)$
\STATE Define valid student depths $\Omega$

\FOR{each training iteration}
    \STATE Sample mini-batch $\mathcal{B}=\{(T_i,S_i,y_i)\}_{i=1}^B$
    \STATE Sample student depth $(E_S,D_S)\sim\mathcal{U}(\Omega)$

    \STATE \textbf{Teacher forward (no gradients):}
    \STATE $\hat{y}^{(T)} \leftarrow f_\theta(T_i,S_i;\,E_T,D_T)$

    \STATE \textbf{Student forward:}
    \STATE $\hat{y}^{(S)} \leftarrow f_\theta(T_i,S_i;\,E_S,D_S)$

    \STATE Compute task loss:
    \STATE $\mathcal{L}_{task} \leftarrow \frac{1}{B}\sum_i \mathcal{L}_{task}(\hat{y}^{(S)}_i,y_i)$

    \STATE Compute distillation loss:
    \STATE $\mathcal{L}_{KD} \leftarrow \frac{1}{B}\sum_i \mathcal{L}_{KD}(\hat{y}^{(S)}_i,\hat{y}^{(T)}_i)$

    \STATE Total loss:
    \STATE $\mathcal{L} \leftarrow \mathcal{L}_{task} + \lambda\,\mathcal{L}_{KD}$

    \STATE Update parameters $\theta \leftarrow \theta - \eta\nabla_\theta\mathcal{L}$
\ENDFOR

\RETURN $\theta^\star \leftarrow \theta$
\end{algorithmic}
\end{algorithm}

Because the original STARK architecture is not designed for early exit, UncL-STARK is fine-tuned to remain predictive at intermediate depths. Each training sample is processed through a full-depth \emph{teacher} path and a randomly truncated \emph{student} path, with knowledge distillation~\cite{hinton2015distilling} transferring supervision from the teacher to all depth configurations [Algorithm~\ref{alg:finetune_teacher_student}].

\subsection{Uncertainty Estimation and Feedback Policy}

UncL-STARK derives a scalar confidence score directly from the corner heatmaps produced by the prediction head. Each heatmap is spatially normalized via softmax, and confidence is computed as the average top-$k$ probability mass across the top-left and bottom-right corner distributions, with higher concentration indicating higher confidence.

Let $\mathbf{H}^{tl}, \mathbf{H}^{br} \in \mathbb{R}^{H \times W}$ denote the predicted corner heatmaps. After softmax,
\[
\mathbf{P}^{c}_{ij}
=
\frac{\exp(\mathbf{H}^{c}_{ij})}{\sum_{u,v}\exp(\mathbf{H}^{c}_{uv})},
\qquad c\in\{\mathrm{tl},\mathrm{br}\}.
\]
Let $\{h^{c}_{(1)},\dots,h^{c}_{(HW)}\}$ be the sorted values of $\mathbf{P}^{c}$.
The confidence score is
\[
\mathcal{C}
=
\frac{1}{2}
\left(
\sum_{m=1}^{k} h^{tl}_{(m)}
+
\sum_{m=1}^{k} h^{br}_{(m)}
\right).
\]
At runtime, confidence computed at frame $t$ determines the depth configuration $(E_{t+1}, D_{t+1})$ for the next frame via a simple threshold-based policy (Algorithm~\ref{alg:adaptive_policy}). This feedback mechanism exploits temporal coherence in video sequences with deeper computation when uncertainty increases.

\begin{algorithm}[t]
\caption{Adaptive Threshold-Based Depth Selection}
\label{alg:adaptive_policy}
\small
\begin{algorithmic}[1]
\REQUIRE Confidence score $\mathcal{C}_t$ at frame $t$
\REQUIRE Thresholds $\tau_{high} > \tau_{low}$
\REQUIRE Depth configurations $(E_{easy},D_{easy})$, $(E_{med},D_{med})$, $(E_{hard},D_{hard})$
\ENSURE Depth selection $(E_{t+1},D_{t+1})$ for next frame

\IF{$\mathcal{C}_t \ge \tau_{high}$}
    \STATE $(E_{t+1},D_{t+1}) \leftarrow (E_{easy},D_{easy})$ \COMMENT{easy frame}
\ELSIF{$\mathcal{C}_t \ge \tau_{low}$}
    \STATE $(E_{t+1},D_{t+1}) \leftarrow (E_{med},D_{med})$ \COMMENT{medium frame}
\ELSE
    \STATE $(E_{t+1},D_{t+1}) \leftarrow (E_{hard},D_{hard})$ \COMMENT{hard frame}
\ENDIF

\RETURN $(E_{t+1},D_{t+1})$
\end{algorithmic}
\end{algorithm}

\section{Experiments and Discussions}
\label{sec:experiments}

We evaluate whether UncL-STARK improves the accuracy--efficiency trade-off of transformer-based tracking by reducing computation through confidence-driven depth adaptation while preserving accuracy. Our experiments proceed in four steps: (i) characterize the accuracy--efficiency curve across depth configurations, (ii) validate heatmap-derived confidence as a proxy for per-frame tracking quality, (iii) evaluate the feedback-driven policy under realistic tracking conditions, and (iv) ablate key components to isolate their contributions.

\subsection{Setup and Evaluation Protocol}
\label{sec:exp_setup}

We fine-tune on the GOT-10k~\cite{huang2021got10k} and LaSOT~\cite{fan2019lasot} training splits and evaluate on the GOT-10k validation and LaSOT test splits. Accuracy is reported using Average Overlap (AO) for GOT-10k and Area Under the Curve (AUC) for LaSOT. When aggregating results across datasets, we report Mean IoU,
$\mathrm{Mean\ IoU} = \frac{1}{S}\sum_{i=1}^{S}\left(\frac{1}{NF_i}\sum_{j=1}^{NF_i}\mathrm{IoU}_{i,j}\right)$,
where $S$ is the number of sequences and $NF_i$ the number of frames in sequence $i$.
Efficiency is measured via GFLOPs reduction, latency improvement, and GPU energy savings.

All models are implemented in PyTorch using the official STARK codebase. We use STARK-S with a ResNet-50~\cite{he2016resnet} backbone pretrained on ImageNet~\cite{deng2009imagenet} and frozen during fine-tuning. The transformer encoder and decoder each contain $N_{\text{enc}}=N_{\text{dec}}=6$ layers. At inference, we evaluate truncated configurations $(E,D)$ with $E,D\in\{1,\ldots,5\}$; layer 0 is always executed to ensure meaningful feature extraction. The corner-based prediction head is unchanged, and confidence is computed directly from its heatmaps.

\subsection{Accuracy--Efficiency Trade-off Across Depth}
\label{sec:tradeoff}

We first analyze how depth affects accuracy and computational cost by evaluating symmetric depth pairs $(1,1),(2,2),(3,3),(4,4),(5,5)$, where $(E,D)=(i,i)$. Symmetric configurations simplify interpretation by scaling encoder and decoder complexity uniformly. We set $\mathrm{MIN}_{\text{enc}}=\mathrm{MIN}_{\text{dec}}=1$, ensuring the first two layers of both encoder and decoder are always executed before any truncation.

\begin{figure}
    \centering
    \includegraphics[width=0.8\linewidth]{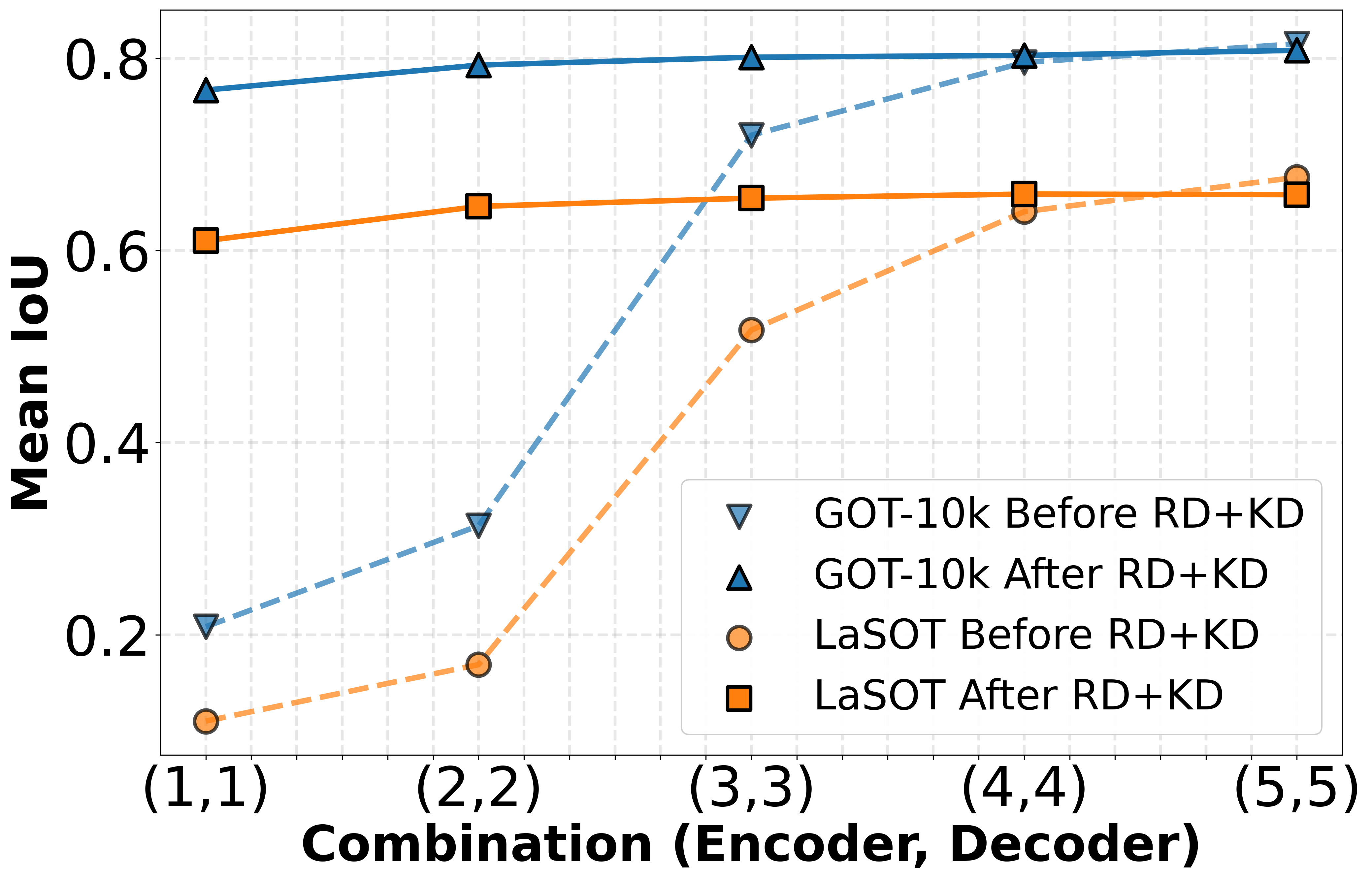}
\caption{Mean IoU versus encoder--decoder depth before and after random-depth (RD) training with knowledge distillation (KD) on GOT-10k (val) and LaSOT (test).}
    \label{fig:combination_eval_copy}
\end{figure}
\begin{table}[h!]
\centering
\caption{Mean IoU performance for different encoder--decoder depth combinations.
Results are reported separately for each dataset and averaged across datasets.
The last column reports the percentage increase in average performance with respect
to the previous depth configuration.}
\label{tab:combination_summary}
\small
\begin{tabular}{c c c c c}
\toprule
\textbf{Combination} &
\makecell{\textbf{GOT-10k}\\\textbf{(val)}} &
\makecell{\textbf{LaSOT}\\\textbf{(test)}} &
\textbf{Average} &
\makecell{\textbf{$\Delta$ Avg}\\\textbf{(\%)}} \\
\midrule
(1,1) & 0.7670 & 0.6102 & 0.6886 & -- \\
(2,2) & 0.7930 & 0.6459 & 0.7195 & +4.48 \\
(3,3) & 0.8012 & 0.6545 & 0.7279 & +1.17 \\
(4,4) & 0.8031 & 0.6587 & 0.7309 & +0.41 \\
(5,5) & \textbf{0.8084} & \textbf{0.6581} & \textbf{0.7333} & +0.33 \\
\bottomrule
\end{tabular}
\end{table}

Figure~\ref{fig:combination_eval_copy} and Table~\ref{tab:combination_summary} report mean IoU for selected depth pairs before and after random-depth fine-tuning with knowledge distillation (RD+KD). Without fine-tuning, shallow configurations exhibit severe accuracy degradation. RD+KD substantially improves all truncated depths, narrowing the gap to the full-depth baseline and making intermediate configurations viable. Configuration $(1,1)$, which executes only the first two encoder and decoder layers (one-third of the full transformer), still incurs noticeable accuracy loss, whereas $(3,3)$ achieves performance close to the full-depth $(5,5)$ configuration.


\begin{table}[t]
\centering
\caption{Adaptive UncL-STARK performance relative to fixed-depth STARK (5,5). Accuracy deltas are relative to baseline; resource values denote percentage reduction.}
\label{tab:overall_compact_full}
\footnotesize
\setlength{\tabcolsep}{5pt}
\begin{tabular}{lcccc}
\toprule
\textbf{Dataset} &
\textbf{AO / AUC} &
$\boldsymbol{\Delta}$ &
\textbf{GFLOPs} &
\textbf{Lat./En.} \\
\midrule
LaSOT (test)    & 65.69 / 64.60 & $-$0.20 / $-$0.17 & \textbf{12.0} & \textbf{8.9 / 10.8} \\
GOT-10k (val)  & 0.8198 / 80.42 & $-$0.19 / $-$0.17 & \textbf{11.9} & \textbf{7.1 / 4.3} \\
GOT-10k (test) & 0.672 / --     & $-$1.04 / --      & \textbf{11.4} & \textbf{7.8 / 6.1} \\
\bottomrule
\end{tabular}
\end{table}

\begin{figure}
    \centering
    \includegraphics[width=0.8\linewidth]{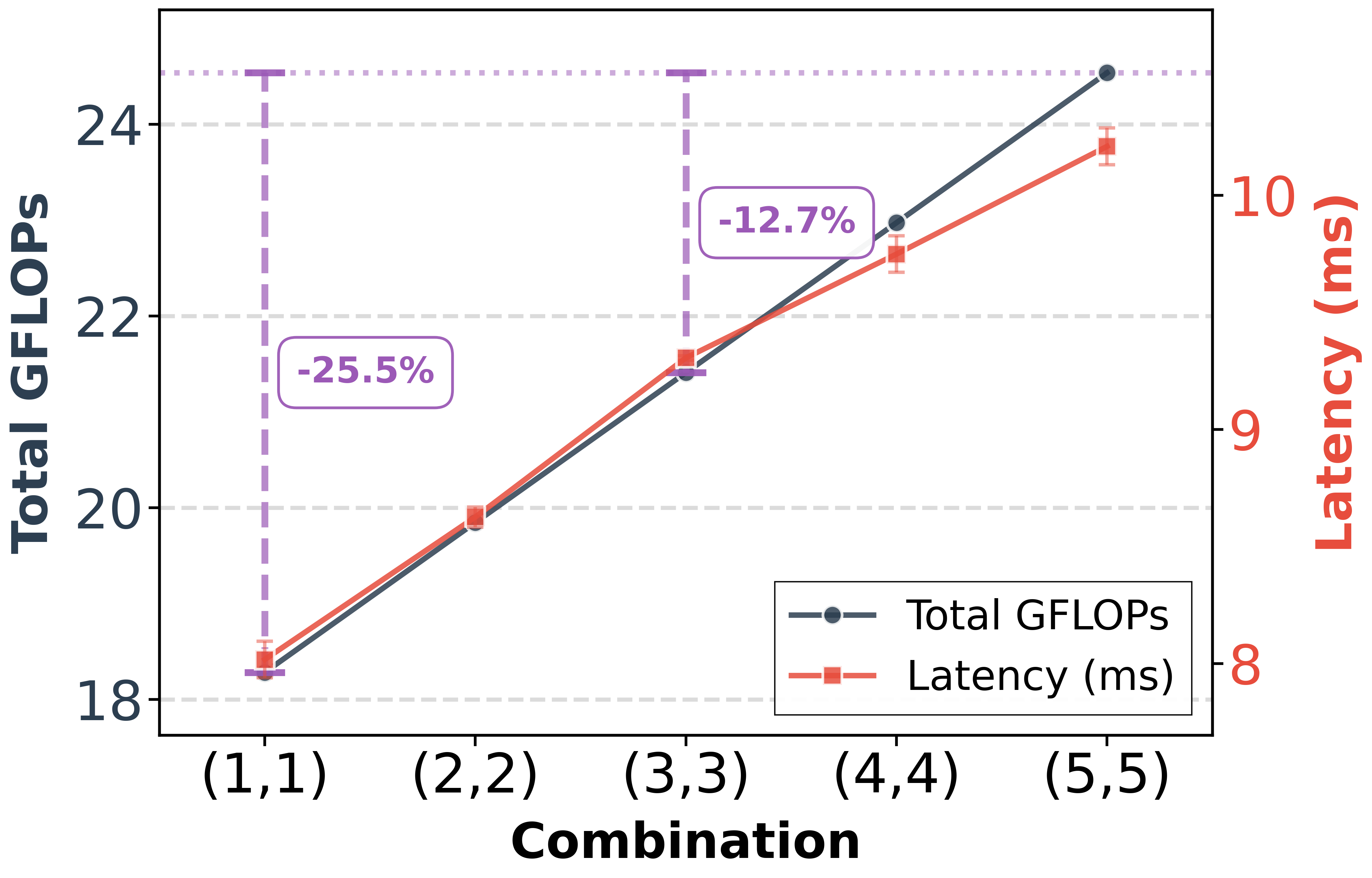}
\caption{Profiled GFLOPs and average latency across depth configurations, both scaling approximately linearly with the number of executed transformer layers.}
    \label{fig:gflops_latency_by_comb}
\end{figure}

Figure~\ref{fig:gflops_latency_by_comb} reports profiled GFLOPs and measured latency across depth configurations, both scaling nearly linearly with the number of executed layers. Configuration $(1,1)$ offers the largest savings (25.5\% GFLOPs reduction) but incurs unacceptable accuracy loss (Figure~\ref{fig:combination_eval_copy}). In contrast, $(3,3)$ achieves 12.75\% GFLOPs savings while maintaining accuracy close to the full-depth baseline, indicating a favorable regime for dynamic depth selection.

\subsection{Validating Heatmap-Based Confidence}
\label{sec:confidence}

UncL-STARK derives a per-frame scalar confidence score directly from the corner heatmaps produced by the prediction head. To validate this choice, we evaluate several alternative heatmap-derived confidence proxies based on their ability to predict tracking quality. Each proxy must satisfy two criteria: (i) strong correlation with per-frame IoU, indicating alignment with tracking performance, and (ii) good calibration~\cite{guo2017calibration,muller2019does}, enabling reliable threshold-based depth selection. All proxies are required to remain computationally lightweight to preserve the efficiency gains of adaptive inference.

\begin{figure}
    \centering
    \includegraphics[width=0.8\linewidth]{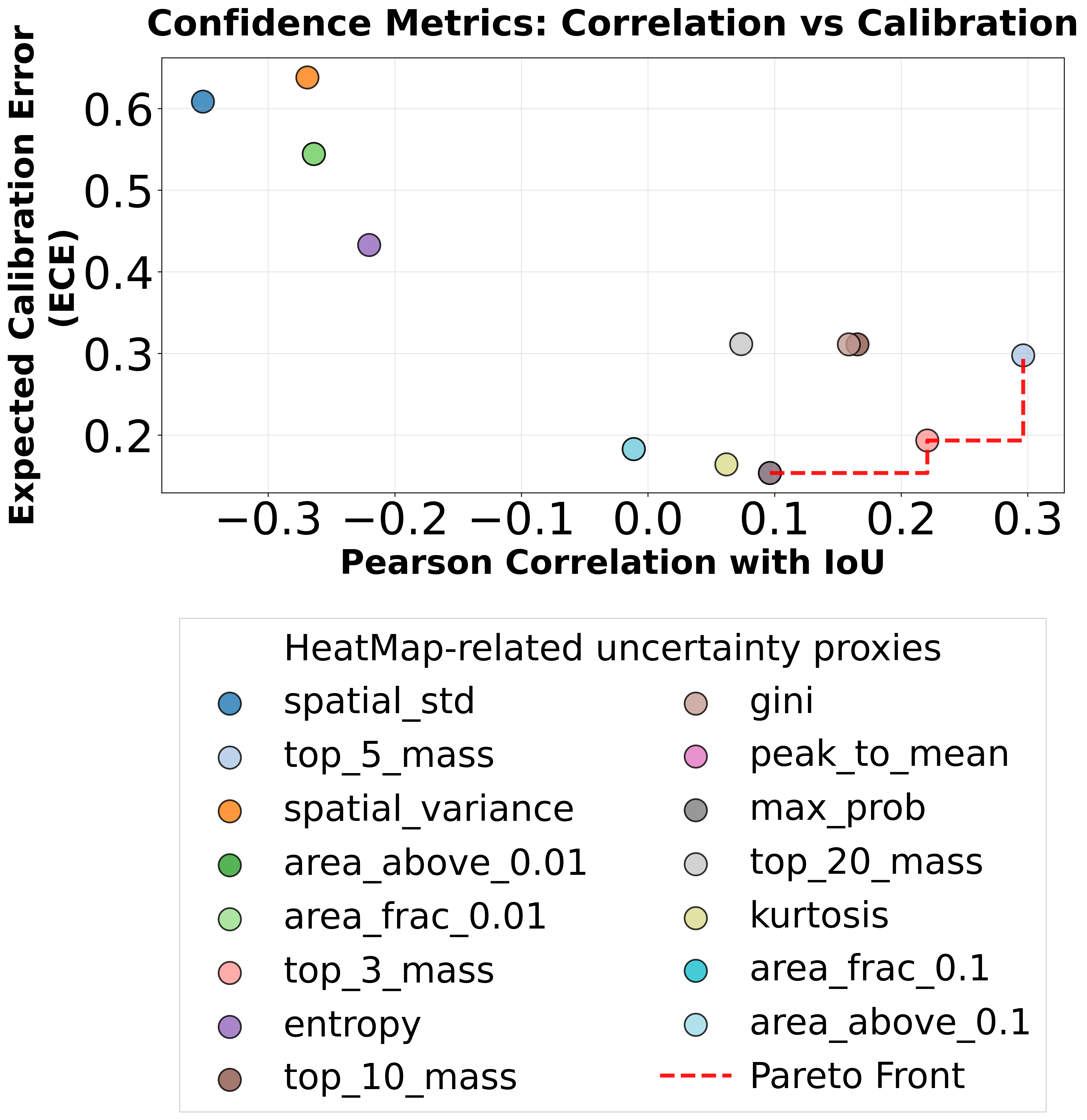}
\caption{Correlation--calibration trade-off of heatmap-derived confidence proxies. Each point is plotted by Pearson correlation with IoU (x-axis) and Expected Calibration Error (ECE, y-axis). Higher absolute correlation and lower ECE indicate better alignment with tracking accuracy. The dashed line denotes the Pareto-optimal frontier; the selected top-$k$ mass estimator lies on this frontier, achieving a favorable balance between correlation and calibration.}
    \label{fig:pearson_vs_ece}
\end{figure}
While some proxies achieve higher correlation with IoU, they often exhibit poor calibration and are therefore unreliable. As shown by the correlation--calibration trade-off in Figure~\ref{fig:pearson_vs_ece}, the top-$k$ mass estimator with $k=3$ lies on the Pareto-optimal frontier, achieving a favorable balance between correlation and calibration (ECE). We therefore select $k=3$, which yields informative confidence distributions and supports robust threshold-based depth selection.

\newcommand{\figscale}{0.90}
\begin{figure*}[t!]
\centering

\begin{minipage}{\figscale\textwidth}
\centering
\includegraphics[width=0.30\textwidth]{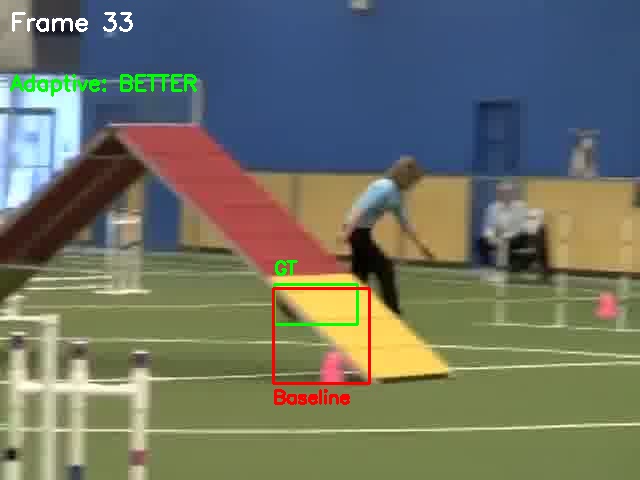}
\includegraphics[width=0.30\textwidth]{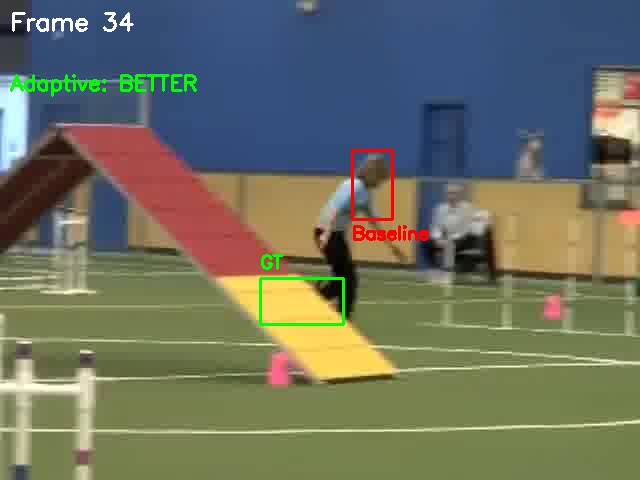}
\includegraphics[width=0.30\textwidth]{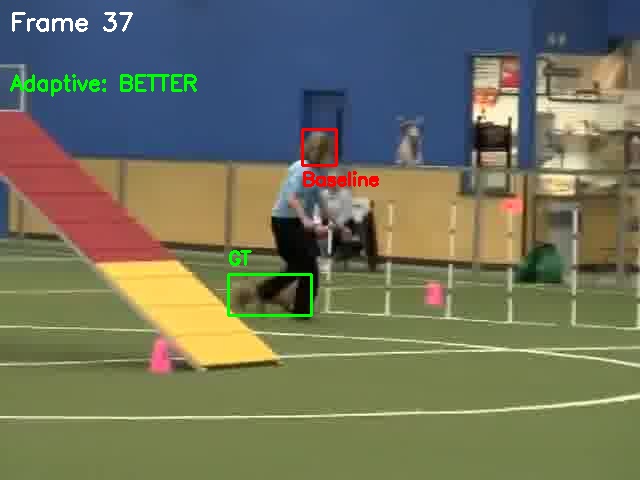}
\captionof{figure}{Baseline tracker predictions at representative frames.}
\end{minipage}

\vspace{4pt}

\begin{minipage}{\figscale\textwidth}
\centering
\includegraphics[width=0.30\textwidth]{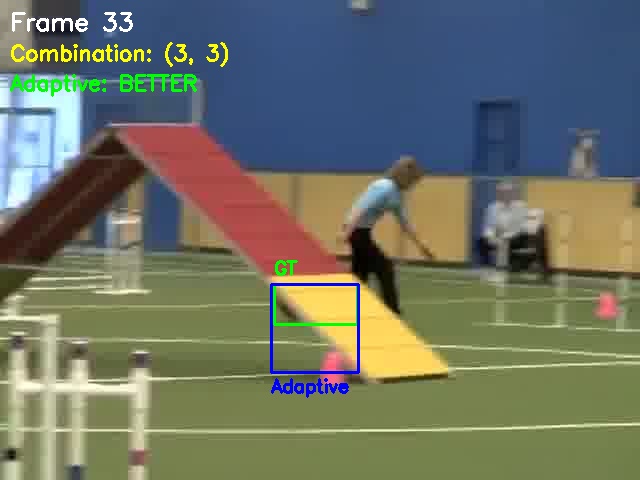}
\includegraphics[width=0.30\textwidth]{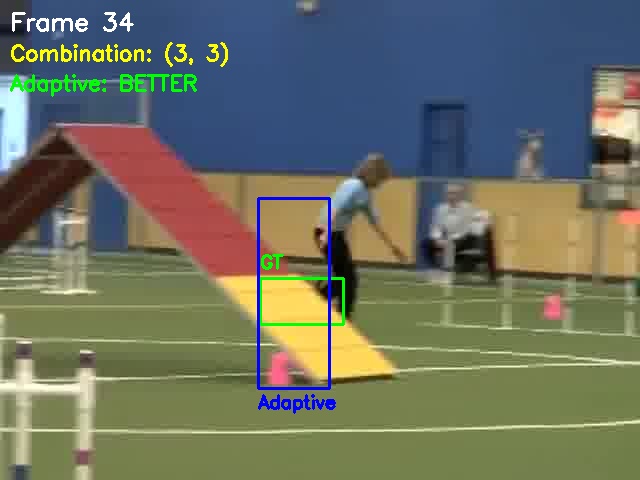}
\includegraphics[width=0.30\textwidth]{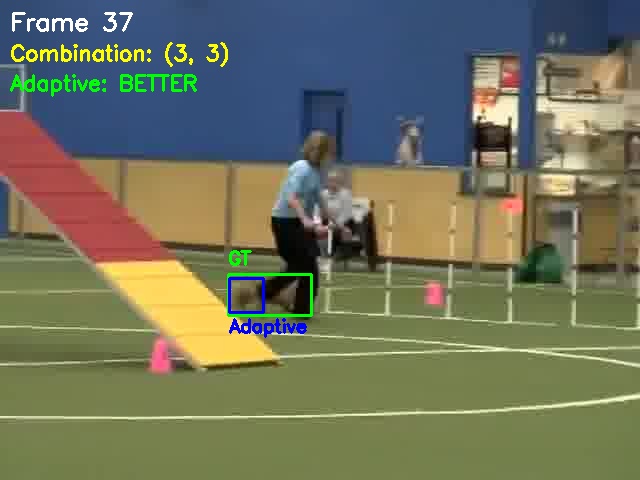}
\captionof{figure}{Adaptive tracker predictions using uncertainty-aware depth selection.}
\end{minipage}

\caption{Qualitative comparison between the fixed-depth baseline and the proposed adaptive tracker on the same video sequence.}
\label{fig:frames_comparison}
\end{figure*}

\subsection{Feedback-Driven Depth Selection}
\label{sec:adaptive}

We evaluate a simple feedback-driven policy that selects among three depth levels using the confidence score computed at frame $t$ to determine the depth for frame $t+1$. The policy considers three configurations:
\[
(2,2)\ \text{(easy)},\quad (3,3)\ \text{(medium)},\quad (5,5)\ \text{(hard)}.
\]
Thresholds are calibrated from the empirical confidence distribution on the validation set. The low and high thresholds are set at the 10th and 85th percentiles, respectively, assigning shallow depth to high-confidence frames, intermediate depth to moderate-confidence frames, and full depth to low-confidence frames. The complete policy is summarized in Algorithm~\ref{alg:adaptive_policy}.

Table~\ref{tab:overall_compact_full} compares the adaptive tracker with the fixed-depth STARK baseline across datasets and evaluation splits. The adaptive policy consistently reduces computation, achieving up to $12.0\%$ GFLOPs savings, $8.9\%$ latency reduction, and $10.8\%$ energy savings, while incurring only marginal accuracy degradation (as low as $0.2\%$). The drop in primary benchmark metrics is limited to $0.17\%$ on LaSOT and $1.04\%$ on GOT-10k (test), indicating that most frames can be processed at reduced depth without substantially affecting accuracy.

These trends are consistent across datasets with different characteristics. Despite LaSOT containing longer and more challenging sequences than GOT-10k, the adaptive policy achieves a similar accuracy--efficiency trade-off, suggesting that confidence-driven depth selection generalizes across sequence lengths and difficulty levels. Overall, UncL-STARK effectively exploits temporal redundancy by allocating higher computation only when uncertainty increases.

\begin{figure}[t]
    \centering
    \includegraphics[width=0.8\linewidth]{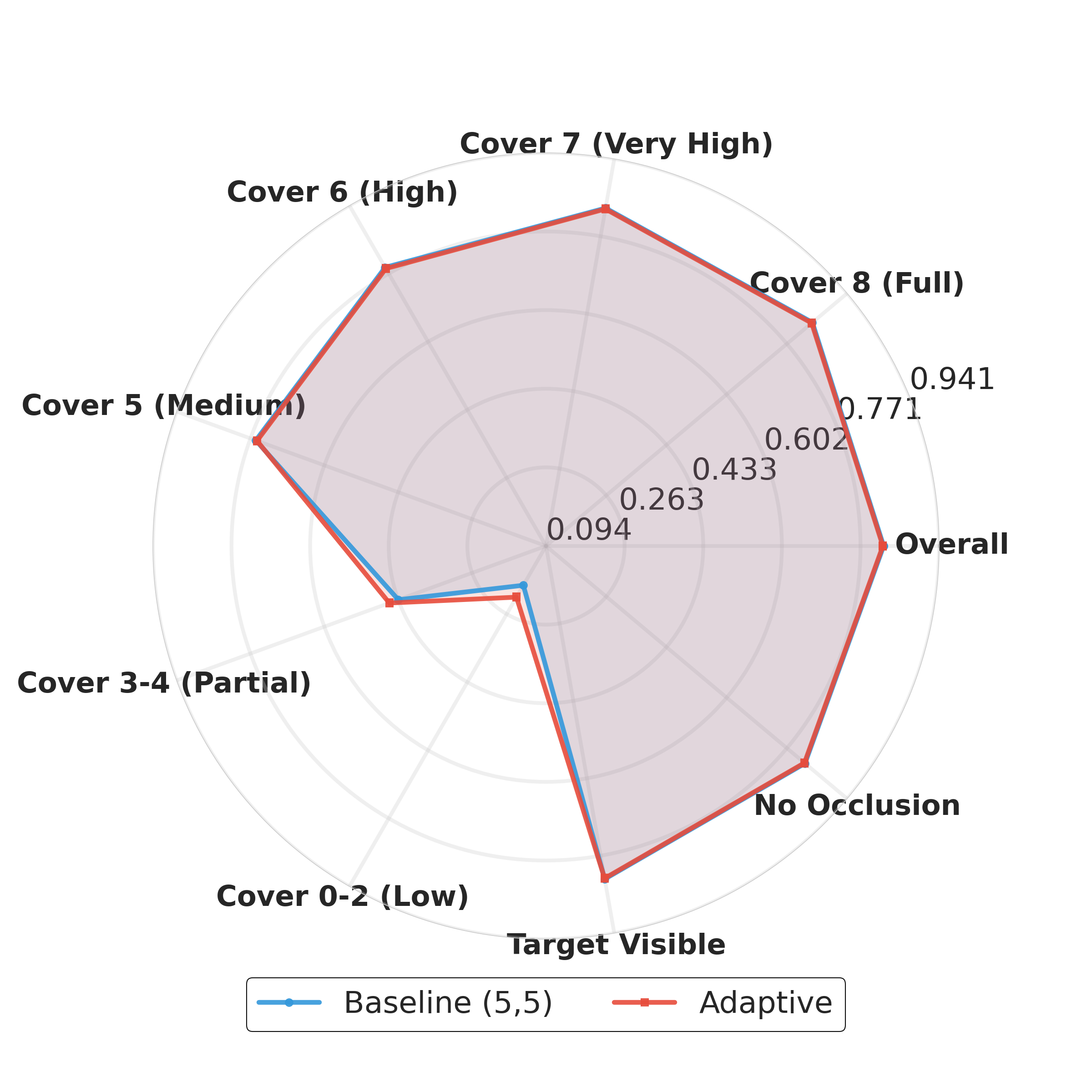}
    \caption{Attribute-based analysis of tracking performance on GOT-10k (val). The attributes represent, in clockwise order, the degrees to which the object is visible (from `Low' to `Full' visibility). The red line represents the adaptive tracking framework's AO, whereas the blue one that of the baseline.}
    \label{fig:spider_got}
\end{figure}

\subsection{Attribute-Based Performance Analysis}

While average performance is favorable, accuracy varies across sequences. To analyze this behavior, we conduct an attribute-based analysis on GOT-10k validation, grouping sequences by object visibility. Figure~\ref{fig:spider_got} shows that the adaptive approach outperforms the full-depth baseline under occlusion (Low and Partial visibility). This counterintuitive behavior is illustrated in Figure~\ref{fig:frames_comparison}, which presents a representative sequence where the adaptive tracker recovers from occlusion more effectively than the baseline.

During occlusion, the adaptive policy assigns shallower depths (e.g., $(3,3)$), producing broader and coarser bounding box predictions. Although less precise, these predictions remain better centered due to the more spatially diffuse features at shallow depths, keeping the search region closer to the true object location. When the object reappears, tracking resumes reliably because coverage of the target area is preserved. In contrast, the full-depth baseline produces tighter predictions during occlusion, which can drift further from the true location; such fine-grained features amplify small errors and hinder recovery. These results indicate that coarser representations can be advantageous under uncertainty, a behavior naturally induced by the adaptive policy without occlusion handling.

\subsection{Ablation Studies}
\label{sec:ablation}

We conduct ablations on GOT-10k validation to isolate the contribution of key components in UncL-STARK.

\begin{table*}[h!]
\centering
\caption{Comparison of baseline (5,5), fixed-depth (3,3), and adaptive tracker variants across LaSOT and GOT-10k. Accuracy is reported using dataset-standard metrics (AUC for LaSOT, AO for GOT-10k). Percentage variations and efficiency gains are computed w.r.t.\ the fixed-depth baseline. Latency refers to measured end-to-end latency; energy is GPU-only (NVML).}
\label{tab:ablation_full_unified}
\begin{tabular}{llcccccc}
\hline
\textbf{Dataset} & \textbf{Method} & \textbf{Acc.} & \textbf{Drop (\%)} & \textbf{FLOPs} & \textbf{Latency} & \textbf{Energy} & \textbf{Adapt. vs. Fixed (3,3)} \\
\hline
\multirow{3}{*}{LaSOT (test)}
 & Baseline (5,5)   & 64.71 (AUC) &  &  &  &  &  \\
 & Fixed (3,3)      & 64.38 (AUC) & 0.50 & 12.7\% & 9.4\% & 11.5\% & \multirow{2}{*}{\cmark} \\
 & Adaptive Tracker & 64.60 (AUC) & 0.17 & 12.0\% & 8.9\% & 10.8\% &  \\
\hline
\multirow{3}{*}{GOT-10k (val)}
 & Baseline (5,5)   & 0.8213 (AO) &  &  &  &  &  \\
 & Fixed (3,3)      & 0.8162 (AO) & 0.62 & 12.6\% & 7.3\% & 4.5\% & \multirow{2}{*}{\cmark} \\
 & Adaptive Tracker & 0.8198 (AO) & 0.19 & 11.9\% & 7.1\% & 4.3\% &  \\
\hline

\hline
\end{tabular}
\end{table*}

\paragraph{Necessity of adaptive depth selection}
A natural question is whether the observed accuracy--efficiency gains can be achieved through static depth truncation. If a fixed-depth configuration could match the adaptive policy, the added complexity of confidence-driven selection would be unnecessary.

We compare the adaptive policy to fixed-depth inference at $(3,3)$, the only static configuration that achieves comparable GFLOPs savings (12.75\%) based on Figures~\ref{fig:combination_eval_copy} and~\ref{fig:gflops_latency_by_comb}. Table~\ref{tab:ablation_full_unified} shows that while static $(3,3)$ attains similar computational savings, it produces substantially worse accuracy, confirming that confidence-driven depth selection is essential for favorable trade-offs. All other static configurations are Pareto-dominated by $(3,3)$, as they either incur larger accuracy drops (shallower depths) or provide smaller savings (deeper depths), making them strictly inferior alternatives.

\begin{table}[t]
\centering
\footnotesize
\setlength{\tabcolsep}{3pt}
\begin{tabular}{l l c c c}
\toprule
\textbf{Dataset} & \textbf{Metric} & \textbf{Random} & \textbf{Threshold} & \textbf{$\Delta$ (\%)} \\
\midrule
\multirow{7}{*}{\textbf{GOT-10k (val)}} 
 & Mean IoU (seq.)        & 0.8015 & 0.8038 & +0.29 \\
 & AO (frames)            & 0.8171 & 0.8198 & +0.33 \\
 & Mean Loss              & 1.8841 & 1.8819 & +0.11 \\
\cmidrule(lr){2-5}
 & GFLOPs (mean)          & 21.98  & 21.61  & \textbf{-1.7} \\
 & Latency (ms)           & 23.09  & 22.91  & \textbf{-1.2} \\
 & Energy (J)             & 234.46 & 237.61 & \textbf{+3.3} \\
\midrule
\multirow{7}{*}{\textbf{LaSOT (test)}} 
 & Mean IoU (seq.)        & 0.6566 & 0.6592 & +0.39 \\
 & AUC (\%)               & 64.50  & 64.60  & +0.15 \\
 & Mean Loss              & 2.0576 & 2.0571 & +0.01 \\
\cmidrule(lr){2-5}
 & GFLOPs (mean)          & 21.88  & 21.60  & \textbf{-1.3} \\
 & Latency (ms)           & 14.19  & 14.06  & \textbf{-0.9} \\
 & Energy (J)             & 3376.91 & 3372.35 & \textbf{-0.1} \\
\bottomrule
\end{tabular}
\caption{Comparison between random depth selection and confidence-threshold-based adaptive depth selection on GOT-10k (val) and LaSOT (test). The threshold-based policy consistently achieves equal or better accuracy while reducing computational cost, latency, or energy consumption, indicating that the adaptive mechanism successfully exploits confidence information to guide depth selection.}
\label{tab:random_vs_threshold_got_lasot}
\end{table}

\paragraph{Effectiveness of random-depth training with knowledge distillation}
Figure~\ref{fig:combination_eval_copy} compares mean IoU across depth configurations before and after random-depth training with knowledge distillation (RD+KD). Without fine-tuning, shallow configurations suffer severe accuracy degradation. RD+KD substantially improves all truncated depths, narrowing the gap to the full-depth baseline and making intermediate configurations viable for adaptive inference. These results validate the necessity of the proposed training strategy for enabling reliable depth truncation in transformer-based tracking.

\paragraph{Alternative confidence proxies}
\begin{figure}[h!]
    \centering
    \includegraphics[width=0.9\linewidth]{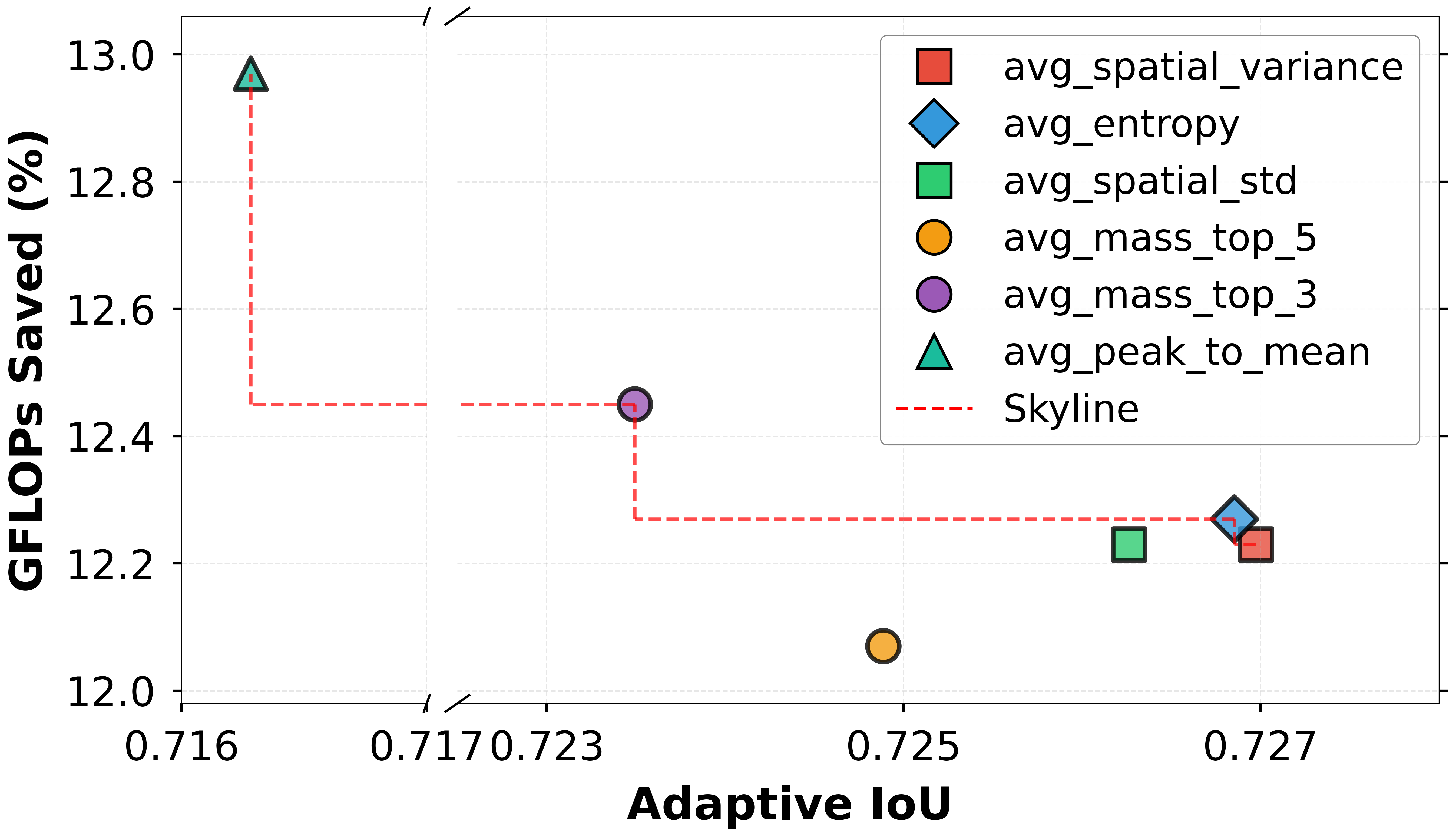}
    \caption{Accuracy--efficiency trade-off of alternative heatmap-derived confidence proxies. Each point represents a different proxy, plotted by the mean IoU achieved by the adaptive tracker (x-axis) and the corresponding GFLOPs savings obtained when the proxy is used to drive depth selection (y-axis). Although some proxies estimate confidence and others uncertainty (with inverted semantics), all serve as scalar indicators of prediction reliability. The dashed line is the Pareto-frontier.}
    \label{fig:conf_ablation_overall}
\end{figure}

We evaluate several heatmap-derived indicators as control signals for adaptive depth selection, each with calibrated thresholds. Figure~\ref{fig:conf_ablation_overall} reports the resulting accuracy--efficiency trade-offs. While some proxies favor accuracy and others computational savings, the average top-$k$ mass with $k=3$ ($avg\_mass\_top\_3$) lies on the Pareto-optimal frontier, achieving a favorable balance between performance and efficiency. We therefore select this proxy as the default confidence measure in UncL-STARK.

\paragraph{Effectiveness of threshold-based policy}
Table~\ref{tab:random_vs_threshold_got_lasot} compares the proposed threshold-based policy with random depth selection, which serves as a non-informative baseline. Under identical architectural and training conditions, the threshold-based policy consistently achieves higher accuracy with comparable or better efficiency across datasets. This confirms that UncL-STARK leverages meaningful confidence signals rather than benefiting from stochastic depth variation.

\section{Conclusion}

We introduced UncL-STARK, an uncertainty-guided inference-time depth adaptation framework for transformer-based visual tracking that reduces computation while preserving accuracy. The method strictly preserves the STARK architecture, enabling safe depth truncation via random-depth training with knowledge distillation and a lightweight uncertainty signal derived from corner heatmaps. A feedback-driven policy couples per-frame confidence to depth selection, exploiting temporal coherence to allocate computation only when uncertainty increases. Experiments on GOT-10k and LaSOT show consistent reductions in GFLOPs, latency, and energy with accuracy remaining within a narrow margin of the full-depth baseline, while ablations confirm that neither static truncation nor random depth selection achieves comparable trade-offs. Notably, under occlusion, shallower depths can yield more stable representations that improve recovery, highlighting uncertainty-guided adaptation as a principled and effective mechanism for efficient, robust tracking within existing transformer architectures.

\bibliographystyle{IEEEtran}
\bibliography{references}
\end{document}